\def\year{2021}\relax
\documentclass[letterpaper]{article} % DO NOT CHANGE THIS
\usepackage{aaai21}  % DO NOT CHANGE THIS
\usepackage{times}  % DO NOT CHANGE THIS
\usepackage{helvet} % DO NOT CHANGE THIS
\usepackage{courier}  % DO NOT CHANGE THIS
\usepackage[hyphens]{url}  % DO NOT CHANGE THIS
\usepackage{graphicx} % DO NOT CHANGE THIS
\usepackage{natbib}  % DO NOT CHANGE THIS AND DO NOT ADD ANY OPTIONS TO IT
\usepackage{caption} % DO NOT CHANGE THIS AND DO NOT ADD ANY OPTIONS TO IT
\usepackage{times}
\usepackage{pbox}
\usepackage{latexsym}
\usepackage{tcolorbox}
\usepackage{multirow}
\usepackage{subfigure}
\usepackage[T1,T2A]{fontenc}
\usepackage[utf8]{inputenc}
\usepackage{tempora}
\usepackage{booktabs}
\usepackage{paralist}
\usepackage{nidanfloat}
\usepackage{xspace}
\usepackage{todonotes}
\usepackage{svg}
\newcommand{\Sref}[1]{\S\ref{#1}}

\newcommand{\Fref}[1]{Figure~\ref{#1}}

\newcommand{\Tref}[1]{Table~\ref{#1}}
\newcommand{\Aref}[1]{Appendix~\ref{#1}}

\newcommand{\ignore}[1]{}

\newcommand\LGBTWikiBio{\textsc{LGBTBio}\xspace}

\newcommand{\markstar}{\textsuperscript{*}}

\usepackage{etoolbox}

\title{Multilingual Contextual Affective Analysis \\of LGBT People Portrayals in Wikipedia}
\author{Chan Young Park\thanks{\hspace{0mm}Equal contribution} \quad Xinru Yan\footnotemark[1] \quad   Anjalie Field\footnotemark[1] \quad  Yulia Tsvetkov\\
   Language Technologies Institute \\
   Carnegie Mellon University\\
   \texttt{\{chanyoun, anjalief, ytsvetko\}@cs.cmu.edu},
   \texttt{xinruyan@alumni.cmu.edu}}

\date{}

\begin{document}
\maketitle
\begin{abstract}
Specific lexical choices in narrative text reflect both the writer's attitudes towards people in the narrative and influence the audience's reactions.
Prior work has examined descriptions of people in English using \emph{contextual affective analysis}, a natural language processing (NLP) technique that seeks to analyze how people are portrayed along dimensions of \emph{power}, \emph{agency}, and \emph{sentiment}.
Our work presents an extension of this methodology to \emph{multilingual} settings, which is enabled by a new corpus that we collect and a new multilingual model.
We additionally show how word connotations differ across languages and cultures, highlighting the difficulty of generalizing existing English datasets and methods.
We then demonstrate the usefulness of our method by analyzing Wikipedia biography pages of members of the LGBT\footnote{Our research focuses on the LGBT sub-community of the LGBTQIA+ community due to data scarcity of other groups.} community across three languages: English, Russian, and Spanish.
Our results show systematic differences in how the LGBT community is portrayed across languages, surfacing cultural differences in narratives and signs of social biases.
Practically, this model can be used to identify Wikipedia articles for further manual analysis---articles that might contain content gaps or an imbalanced representation of particular social groups.

\end{abstract}

\section{Introduction}
\label{sec:Intro}
\begin{figure}
\centering
\includegraphics[width=0.95\columnwidth]{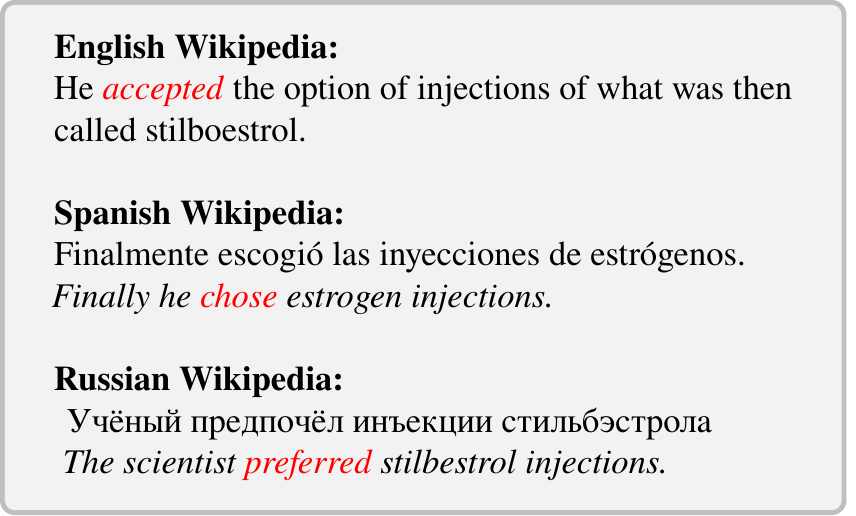}
\caption{Example from Alan Turing's biography page on Wikipedia in different languages. Verb choice in different languages can have subtly different connotations.}
\label{fig:intro_example}
\end{figure}

In 1952, Alan Turing was prosecuted for being gay and subsequently underwent a hormonal injection; two years later he committed suicide. \Fref{fig:intro_example} shows parallel sentences drawn from his English, Spanish, and Russian Wikipedia pages. Although all three sentences describe the same situation, their connotations subtly differ. The English edition uses the verb \textit{accepted}, which suggests that Turing had little control over the situation (low agency). In contrast, the verbs \textit{chose} in Spanish and \textit{preferred} in Russian imply that he actively made the decision (high agency). The verb \textit{preferred} in Russian can even imply positive sentiment towards the injections, while the English connotation is more negative. Thus, Russian, Spanish, and English readers who search for Alan Turing on Wikipedia may form different impressions about this part in his life.

These subtle differences in phrasing can be indicative of social norms and perceptions about social roles \cite{eckert2000language,tannen1994gender}. In general, analyzing narratives about people sheds light on stereotypes and power structures \cite{Hall:Braunwald:1981,fournier2002social} and examining how these concepts differ across cultures is an important component of social-oriented analysis \citep{Almeida:2009,Balsam:2011,men:2013}.
In the example in \Fref{fig:intro_example}, these discrepancies in phrasing could indicate that stereotypes and bias about LGBT people manifest differently in Russian, English, and Spanish-speaking cultures. On Wikipedia, manifestations of stereotypes are violations of the platform's ``Neutral Point of View'' policy.\footnote{\url{https://en.wikipedia.org/wiki/Wikipedia:Neutral_point_of_view}} 

In this work, we develop computational methods that facilitate large-scale analyses of people described in multilingual narrative text. This technology can aid readers or writers, such as Wikipedia editors or journalists, in identifying biases in sets of articles that can be further analyzed to understand social stereotypes or edited in order to reduce bias.

Recent advances in NLP have analyzed stereotypes and biases in narratives in English  \citep{bamman-etal-2013-learning,wagner2015s,sap-etal-2017-connotation}. \citet{field2019contextual} establish a framework called Contextual Affective Analysis (CAA) that focuses on affective dimensions of \emph{power} (strength/weakness), \emph{agency} (activeness/passiveness), and \emph{sentiment} (goodness/badness).  Analyzing portrayals of people along these dimensions (e.g., are men or women portrayed as more powerful?) has revealed stereotypes and bias in various domains, including movie scripts and newspaper articles \citep{rashkin-etal-2016-connotation,sap-etal-2017-connotation,field-tsvetkov-2019-entity,field2019contextual,antoniak2019narrative}. 

Measuring these affective dimensions relies on \emph{connotation frames}---lexicons of verbs annotated to elicit implications \citep{rashkin-etal-2016-connotation,sap-etal-2017-connotation}.
These connotations can be subtle, and the same verb often has different connotations in different contexts \cite{field2019contextual,field-tsvetkov-2019-entity}.
Until now, these manually-annotated affective lexicons have existed only for English.
While lexicons can be machine-translated \citep{rashkin-etal-2017-multilingual,mohammad-2018-obtaining}, no work has yet conducted in-language evaluations of connotation translatability nor attempted extensive analysis in other languages.

Our ultimate goal is to measure the power, agency, and sentiment of people described in multilingual text; we here focus on English, Spanish, and Russian. These measurements allow us to conduct both in-language analysis (in Russian text, are LGBT people portrayed as more powerful than non-LGBT people?) and cross-language analysis (is the power differential between LGBT people and non-LGBT people greater in English or in Russian?). To accomplish this, we first crowdsource annotations of connotation frames in English, Spanish, and Russian (\Sref{sec:data}). We then analyze how connotations vary across contexts and languages in these data in order to demonstrate why existing English data sets are insufficient (\Sref{sec:connanalyze}).  
With the new dataset, we develop multilingual CAA classifiers of power, agency, and sentiment (\Sref{sec:model}).

Finally, we demonstrate the usefulness of our methodology in a semi-automated analysis \Sref{sec:LGBT}, by collecting a new corpus (\LGBTWikiBio) and analyzing how members of the LGBT community are portrayed on Wikipedia in different languages. 
Our results show that the biography pages of LGBT people in our corpus typically contain more negative connotations than the pages of other people in Russian. In contrast, pages about the same LGBT people are more positive or neutral compared to other pages in English, and generally neutral or mixed in Spanish. These trends align with survey results about perceptions of LGBT people in English, Spanish, and Russian-speaking countries \citep{Flores2019}.

The key contributions of our work are annotated datasets, machine learning models, and a general methodology that enables nuanced analyses of narratives about people across languages. We additionally present an analysis of LGBT people on Wikipedia, whereas extensive prior computational work on Wikipedia biography pages has focused primarily on male/female gender bias \citep{callahan2011cultural,recasens-etal-2013-linguistic,wagner2015s,Chandrasekharan:2017}.\footnote{Code and data are publicly available at \url{https://github.com/chan0park/multilingual-affective-analysis}.}

\section{Crowdsourcing Contextualized Connotation Frames}
\label{sec:data}
We first collected a corpus of multilingual connotation frames in English, Spanish, and Russian. Connotation frame annotations ask annotators to answer questions about the power, sentiment, and agency of the agent (approximated as the grammatical subject) and theme (approximated as the object) of verbs.
At a high level, we seek to answer: (1) Does the subject have more/less/equal \textbf{power} as the object? (2) Does the subject have low/moderate/high \textbf{agency}? (3) Does the writer feel positive/negative/neutral about the subject (Sent\textsubscript{subj})? (4) Does the writer feel positive/negative/neutral about the object (Sent\textsubscript{obj})?

Consider the sentence: \textit{The firefighter rescued the boy.}
The verb \textit{rescued} implies that the subject, \textit{firefighter} has more power than the object, \textit{boy}. The firefighter is also active and in control of his actions, which shows high agency. Because \textit{rescuing} is a positive action (e.g. as opposed to \textit{killing}), the writer likely feels positively about the subject. In the absence of information conveying positive or negative sentiment about \textit{boy},  we can infer that the writer feels neutrally about the object. 

Our connotation frames differ from existing lexicons in two primary ways: first, no prior work has collected connotation frames in languages other than English, and second, we collect all annotations in complete contexts drawn from newspaper articles, e.g., \textit{the firefighter rescued the boy}, whereas prior work uses either simplified tuples or artificial placeholders, e.g., \textit{X rescues Y}  \citep{sap-etal-2017-connotation,rashkin-etal-2016-connotation}.
Because these affective dimensions can be difficult to define, we took numerous steps to ensure annotations would be of high quality.
We briefly summarize here and provide details in \Aref{sec:schema} to facilitate reproducibility.
 
We first extracted frequent verbs and contexts that are representative of each verb's most common usage, and then asked annotators to annotate verbs in these contexts.
For each language, we extracted all (subject, object, verb) tuples from a \textit{News Crawl} corpus.\footnote{A large monolingual corpus of newspaper articles %, from WMT19 
\citep{barrault-EtAl:2019:WMT}. Throughout this work, parsing was done with SpaCy.}
We chose the $300$ most frequent transitive verbs to annotate. For each verb, we took the three most common (subject, object, verb) tuples as the most representative context. 
We restricted tuples to have at least one human subject or object by using the list of words in \textit{noun.person} category of WordNet  \citep{fellbaum2012wordnet}.\footnote{Extensive list of English nouns denoting people \citep{Miller:1995:WLD:219717.219748}; we translated to other languages with Google Translate.}
We then pulled phrases containing the chosen tuples from the news corpus, which served as our data to be annotated.

We used the same interfaces as \citet{rashkin-etal-2016-connotation} and \citet{sap-etal-2017-connotation}, with minor modifications based on feedback during pilot studies.\footnote{We will provide full annotation interface in the released data.}
For non-English annotation tasks, a native speaker translated the task instructions into the target language.
We restricted the pool of annotators to the United States for English, Russia for Russian, and to several South American countries for Spanish.
For each of the three target languages, we collected power, agency, and sentiment annotations for $300$ verbs in three contexts each ($900$ instances). For each instance, we collected judgements from three annotators, leading to $32,400$ total annotations.

Despite the steps taken to ensure annotation quality, we suspect that some annotators paid more attention to the task instructions and generated higher-quality judgements than others. We correct for this by discarding annotations from 14.4\% of workers who frequently disagreed with other annotators (\Aref{sec:schema}). Krippendorff's alpha, averaged across tasks, was 0.22 for English, 0.31 for  Russian, and 0.26 for Spanish. These agreement scores are comparable to prior work \citep{rashkin-etal-2016-connotation,sap-etal-2017-connotation} and reflect the subjective nature of connotation frames; very high agreement would suggest that we over-simplified the task, for example by choosing non-representative samples to annotate. Additionally, the most common case of annotator disagreement was when one annotator labeled an instance as neutral and another did not, meaning polar-opposite annotations were rare; if we only count polar-opposite annotations as disagreements, the average pairwise agreement is 92.5\%.
We describe additional restrictions used to ensure annotation quality and full agreement metrics in \Aref{sec:schema}.

To aggregate annotations, we mapped each judgement to a $(-1, 0, 1)$ value and averaged annotator scores.
We then ternerized the aggregated scores by labeling connotations as positive $[1, 0.35]$, neutral $(0.35, -0.35)$, and negative $[-0.35, -1]$. With these boundaries, a connotation is only scored as positive or negative if at least two annotators labeled it with this polarity, and thus, samples where annotators disagreed were labeled as neutral---not clearly indicative of a positive or negative connotation.

\section{Crowdsourced Connotation Analysis}
\label{sec:connanalyze}
As described in \Sref{sec:data}, our data differs from existing lexicons in two primary ways: (1) we collected annotations in context and (2) in various languages. We analyzed our data to assess how these differences impact lexicon quality.

\begin{table}
    \centering
        \begin{tabular}{cccc}
    & English & Russian & Spanish \\ \hline \hline
    \textbf{Power} & $\downarrow$29.2\% & $\downarrow$24.5\% & $\downarrow$26.7\%  \\
    \textbf{Agency} & $\downarrow$31.1\% & $\downarrow$30.9\% & $\downarrow$35.2\% \\
    \textbf{Sent(subj)} & $\downarrow$18.5\% & $\downarrow$18.4\% & $\downarrow$29.6\% \\
    \textbf{Sent(obj)} & $\downarrow$20.2\% & $\downarrow$23.6\% & $\downarrow$29.8\% \\
    \end{tabular}
    
    \caption{Assessment of how much information is lost when using verb-level (``rescues'') annotations instead of context-level (``the firefighter rescued the boy''). Accuracy decreases by nearly 20\% for all languages.}
    \label{tab:context2}
\end{table}

\paragraph{How important is contextualization?}
\label{sec:context}
We first address this question by examining how much accuracy would be lost if we use a single connotation score for each verb (e.g., one score for \textit{deserves} in all contexts where it appears), rather than different scores when the verb appears in different contexts (e.g., different scores for \textit{the boy deserves a reward} and \textit{the boy deserves a punishment}) \citep{field-tsvetkov-2019-entity}. To compute this, we ``decontextualize'' our lexicons by averaging annotations for each verb, regardless of the context it was annotated in, into a single verb-level score. We compare this decontextualized score with the contextualized score in our actual dataset, where we only average annotations for verbs annotated in the same context. \Tref{tab:context2} shows how often the verb-level score differs from the context-level score. If verbs had the same connotations in different contexts, all values in this table would be 0\%. Instead, we see that ignoring contextualization can result in an $>30\%$ drop in accuracy. While \citet{field-tsvetkov-2019-entity} have similar findings for sentiment connotations in English, \Tref{tab:context2} extends these findings to power and agency connotations and to Spanish and Russian.

\begin{table}
    \centering
        \begin{tabular}{cccc}
    &  Russian & Spanish \\ \hline \hline
    \textbf{Power} & $\downarrow$37.6\% & $\downarrow$51.1\%  \\
    \textbf{Agency} & $\downarrow$51.2\% & $\downarrow$47.8\% \\
    \textbf{Sent(subj)} & $\downarrow$21.6\% & $\downarrow$34.8\% \\
    \textbf{Sent(obj)} & $\downarrow$30.4\% &  $\downarrow$37.0\% \\
    \end{tabular}
    
    \caption{Assessment of information lost when using translated annotations instead of in-language annotations, evaluated over $125$ Russian and $135$ Spanish verbs that overlapped with English annotations.}
    \label{tab:translation}
\end{table}

\paragraph{How important are in-language annotations?}
\label{sec:LGBTCorpnguage}
We cannot directly compare contextualized annotations across languages because we annotate different contexts for different languages; however, there is some overlap between the most frequent verbs in each language. For each non-English language, we first aggregate annotations into the same decontextualized verb-level scores as in the previous paragraph. We then use Google Translate to translate verbs into English and intersect them with our annotated English verbs.\footnote{Google Translate has previously been used to obtain multilingual lexicons \citep{Mohammad13}
} We discard any translation pairs that native Russian and Spanish speakers judged to be inaccurate or questionable (11 in Russian; 9 in Spanish). Finally, we measure how often the decontextualized English annotations differ from the original in-language annotations.

\Tref{tab:translation} reports the results. Because we assess the decontextualized scores, if word-level translation were effective for obtaining multilingual connotations, all scores would be similar to the ones in \Tref{tab:context2}. Instead, they are substantially higher, showing that information is lost because of translation.  Both Tables \ref{tab:context2} and \ref{tab:translation} suggest that our annotations can facilitate higher-quality analyses than ones collected in prior work. Because word-level translations are often inaccurate, in \Sref{sec:eval}, we also explore using a cross-lingual model and sentence-level translation to translate connotations.

\section{Classification of Connotations}
\label{sec:model}
Our goal is to develop a methodology for analyzing how people are portrayed in different languages.
The multilingual annotations alone are insufficient, as 900 contexts represent a tiny subset of all verb usages in a corpus.
Thus, we need a method to obtain connotation scores for unseen verbs and contexts. We describe our method here and provide reproducibility details in \Aref{sec:appendix_classifier}.

We follow prior work in developing a supervised classifier trained on our contextualized multilingual annotations that can predict a connotation frame label $y_v$ for any unseen (in-context) verb $v$ \citep{rashkin-etal-2016-connotation,field2019contextual}.
Unlike prior models, ours is trained on contextualized annotations, and it leverages pre-trained cross-lingual language models (CLMs). CLMs produce language-agnostic feature representations, allowing us to combine different languages in the training and test data.

\begin{table}[t]
\centering
\begin{tabular}{l||lllll}
\toprule
    Tgt &  Src &    Sent\textsubscript{subj} &  Sent\textsubscript{obj}  &  Pow. & Agen. \\
    \midrule
 \multirow{3}{*}{EN} &   EN &  \textbf{43.4}\markstar &  43.0 &  \textbf{41.1} &  \textbf{48.2}\markstar  \\
  &   ES &  38.1 &  43.4 &  29.5 &  43.4 \\
  &   RU &  41.1 &  \textbf{44.3} &  40.1 &  41.4 \\
 \midrule
 \multirow{3}{*}{ES} &   EN &  38.9 &  36.6 &  24.5 & 31.3 \\
  &   ES &  \textbf{49.5}\markstar &  \textbf{51.2}\markstar &  \textbf{43.6}\markstar &  \textbf{43.6}\markstar \\
  &   RU &  39.0 &  42.2 &   34.0 &   38.9  \\
 \midrule
 \multirow{3}{*}{RU} &   EN &  43.6 &   49.2 &  36.4 &  44.5 \\
  &   ES &  37.2 &   49.3 &  38.2 &  42.7 \\
  &   RU &  \textbf{46.4}\markstar &  \textbf{54.9}\markstar &  \textbf{45.3}\markstar &  \textbf{49.9}\markstar \\
\bottomrule
\end{tabular}
\caption{Macro F1 score of classifiers trained and evaluated with different target and source languages. Matching the language of the training and test data achieves better results than training on different languages. 
Asterisks are added to the best performing model in each (language, attribute) pair when it is significantly better than the second-best model (paired t-test, $*$: $p{<}0.05$). 
} 
\label{tab:model_res}
\end{table}

We obtain multilingual embedding representations ($c_v$) of verbs in-context by extracting the last hidden layer of the pretrained model called XLM, which achieves state-of-the-art performance in a variety of cross-lingual tasks \citep{conneau2019cross}. We then use $c_v$ as features in a classifier.

Our primary classifier is a logistic regression model with sample weighting. 
The classifier is trained to predict a connotation frame label $y_v$ from the input representation $c_v$.
Sample weights are tuned over a dev set.
While the classifier architecture is the same as in \citep{rashkin-etal-2016-connotation,field2019contextual}, inputs to our model differ.
Prior connotation frame lexicons only contain verb-level annotations, and classifiers were trained using non-contextual embeddings (e.g., word2vec \citep{rashkin-etal-2016-connotation}) or decontextualized embeddings (e.g., ELMo \citep{field2019contextual}). With new annotations over verbs in-context, we directly train the classifier on contextual embeddings. Finally, we use the trained classifier to predict connotation frame labels for verbs provided with their context in a target language corpus.

\section{Connotation Classification Evaluation}
\label{sec:eval}

We evaluate our model on the contextualized multilingual annotation data described in \Sref{sec:data} using $5$-fold cross-validation and splitting data into train, development, and test sets with the ratio of $6$:$2$:$2$.
\Tref{tab:model_res} reports macro F1 scores\footnote{We provide evidence that our model performs comparably with prior work in \Aref{sec:appendix_classifier}} for single-language evaluations, where we train on the training set of one language (``Src'') and evaluate on the test set of a second language (``Tgt'').
Unsurprisingly, training and testing on the same language achieves better performance than training on one language and testing on another.
While the cross-lingual model works well in certain cases, in most cases transferring languages yields a substantial decrease in performance. For power in Spanish, F1 decreases by $19$ points ($44$\%) when the training language is English instead of Spanish. These results offer further evidence of the importance of in-language connotations.

\begin{table}
    \centering
    \begin{tabular}{c||cccc}
    \toprule
         Tgt  & Sent\textsubscript{subj} & Sent\textsubscript{obj} & Power & Agency \\
         \midrule
         ES & 21.1 & 20.1 & 31.7 & 27.3 \\
         RU & 18.9 & 30.8 & 34.4 & 24.2 \\
         \bottomrule
    \end{tabular}
    \caption{Macro F1 for Machine Translation approach is strictly worse than for cross-lingual model (\Tref{tab:model_res}).}
    \label{tab:eval_mt}
\end{table}

\paragraph{Machine Translated Classification}
In \Tref{tab:eval_mt}, we consider an alternative approach to in-language annotations. We translate the Spanish and Russian test sentences into English through Google Translate, and then we use the model trained on English annotations to predict connotations in these translated sentences. This method simulates translating a corpus into English, and using a model trained on English for analysis. Machine translation performs strictly worse than a model training on in-language data, suggesting it cannot replace in-language data.

\begin{table}[t]
\centering
\begin{tabular}{l||lllll}
\toprule
    Tgt &  Src &    S\textsubscript{subj} &  S\textsubscript{obj}  &  Pow. & Agen. \\
    \midrule
 \multirow{3}{*}{EN} &   EN &  43.4 &  43.0 &  41.1 &  48.2  \\
  &   +ES &  44.8  & \textbf{45.2}\markstar  & 40.5 & 49.7 \\
  &   +RU &  \textbf{46.5}\markstar & 43.2 & \textbf{41.8} & 49.9 \\
  &   +ES+RU &  45.0 & 44.3 & 41.7 & \textbf{50.0}\markstar \\
 \midrule
 \multirow{3}{*}{ES} &  ES &  49.5 &  51.2 &  \textbf{43.6} &  43.6 \\
  &  +EN &  50.4 & 51.6  & 36.4 & 45.5 \\
  &  +RU & 51.0  & \textbf{55.0}\markstar &  {42.1} &  \textbf{45.6}\markstar \\
  &  +EN+RU &  \textbf{51.8}\markstar & 54.8 &  40.8 &  44.9 \\
 \midrule
 \multirow{3}{*}{RU} & RU &  46.4 &  54.9 &  45.3 &  49.9 \\
    & +EN & 45.6 & 55.7 &  44.1 &  50.9\\
    & +ES &  46.0 & \textbf{59.2}\markstar &  42.1 &  49.8 \\
    & +EN+ES &  \textbf{47.7}  & 53.7 &  \textbf{46.9}\markstar  &  \textbf{51.7}\markstar\\
\bottomrule
\end{tabular}
\caption{Macro F1 score of classifiers, where in-language training data is augmented with training data from other languages (e.g., Tgt=EN, Src=+ES indicates the model was trained on English and Spanish data). Augmentation improves performance in most cases.
Asterisks are added to the best performing model in each (language, attribute) pair if there is a significant improvement made by data augmentation (paired t-test, $*$: $p{<}0.05$).
} 
\label{tab:model_augment}
\end{table}

\paragraph{Augmented Cross-Lingual Connotation Classification}
While \Tref{tab:model_res} suggests that in-language training data results in better performance than out-of-language training data, we also explore using in-language and out-of-language data in combination. This experimentation is based on the hypothesis that while connotations differ in different languages, there may be enough overlap for the cross-lingual model to learn useful signals from out-of-language data. \Tref{tab:model_augment} shows results. For all connotation dimensions, the best performing augmented models outperform the non-augmented models. In \Sref{sec:LGBT}, we use the best performing models from \Tref{tab:model_augment} to analyze biography pages of LGBT people.

\section{Case Study: Multilingual Affective Analysis of LGBT People}
\label{sec:LGBT}

Finally, we demonstrate how the new corpus of annotations (\Sref{sec:data}) and the cross-lingual model (\Sref{sec:model}) facilitate multilingual analysis by examining portrayals of LGBT people on Wikipedia. We focus on LGBT people because discrimination against the LGBT community is an increasingly important global issue.
Although pride marches are held $\ge158$ cities world-wide \cite{lisitza_2017}, social oppression is prevalent in many countries \citep{Balsam:2011,doi_stewart_2019}.
Nevertheless, little prior computational work has studied narratives about the community, likely due to data scarcity \citep{mendelsohn2020framework}. To facilitate analysis, we collect a new corpus, titled \LGBTWikiBio, which contains Wikipedia biography pages of LGBT people and pages of non-LGBT people (controls). We explain the motivation and methods behind our corpus collection in \Sref{sec:LGBTCorp} and discuss results in \Sref{sec:analysis}.\footnote{All of our analysis is conducted over publicly available data. We do not expect our results to have any negative effects on the individuals analyzed or the broader LGBT community.}

\subsection{\LGBTWikiBio Corpus}
\label{sec:LGBTCorp}

We collected a multilingual corpus of $1,340$ Wikipedia biography pages for people in the LGBT community using Wikidata properties and lists of LGBT people from Wikipedia (details in \Aref{sec:appendix_matching}).
This corpus allows us to analyze how the same person is portrayed in different languages. However, we cannot draw conclusions from this corpus alone, because we need to control for overall language differences. For example, if we find that LGBT people have higher power in English pages than in Russian pages, we cannot determine if this difference occurs because LGBT people are actually described differently or because English verbs tend to have more positive power connotations than Russian verbs.

To isolate the effects of our variable of interest (i.e., sexual orientation), we built a control corpus by matching each LGBT person with a non-LGBT person who has similar characteristics using a matching method from \citet{anonymous20matching}. Our goal is not to construct perfectly-matched pairs, and we do not analyze any single pair. Rather, we seek to build a control corpus that has a similar distribution of characteristics as the main corpus \textit{on aggregate}, excepting of sexual orientation, which is a common practice in causal inference studies \citep{Stuart2010}.

To identify matches, for each person $Y$ in the LGBT corpus, we first scraped their associated Wikipedia categories (e.g., \textit{American fashion businesspeople}). 
We then identified all other people listed in these categories and constructed weighted TF-IDF vectors from each person's associated categories. We selected the control person as the one who has the most similar category vector as $Y$. We refer to \Aref{sec:appendix_matching} and \citet{anonymous20matching} for details about the corpus and matching algorithm. The $1,340$ pages about LGBT people and their matched controls together constitute the  corpus.

\subsection{Contextual Affective Analysis of Narratives Describing LGBT People}
\label{sec:analysis}
We first use our model to compute high-level trends that identify directions for further exploration and then leverage our model to extract samples that exemplify these trends.
Given the difficulty of the task and the subjective nature of connotations, we do not recommend using our model to draw black-box inferences, but rather suggest it is most useful in facilitating manual analyses that would otherwise be prohibitively expensive given the volume of Web-scale data. We also note that Wikipedia is constantly changing, and we observed several articles in our data set change over the course of this work. Thus, our results may not generalize to data collected at other times or to other articles beyond the specific ones included in our corpus.

\begin{figure}[t]
    \centering
    \includegraphics[width=\columnwidth]{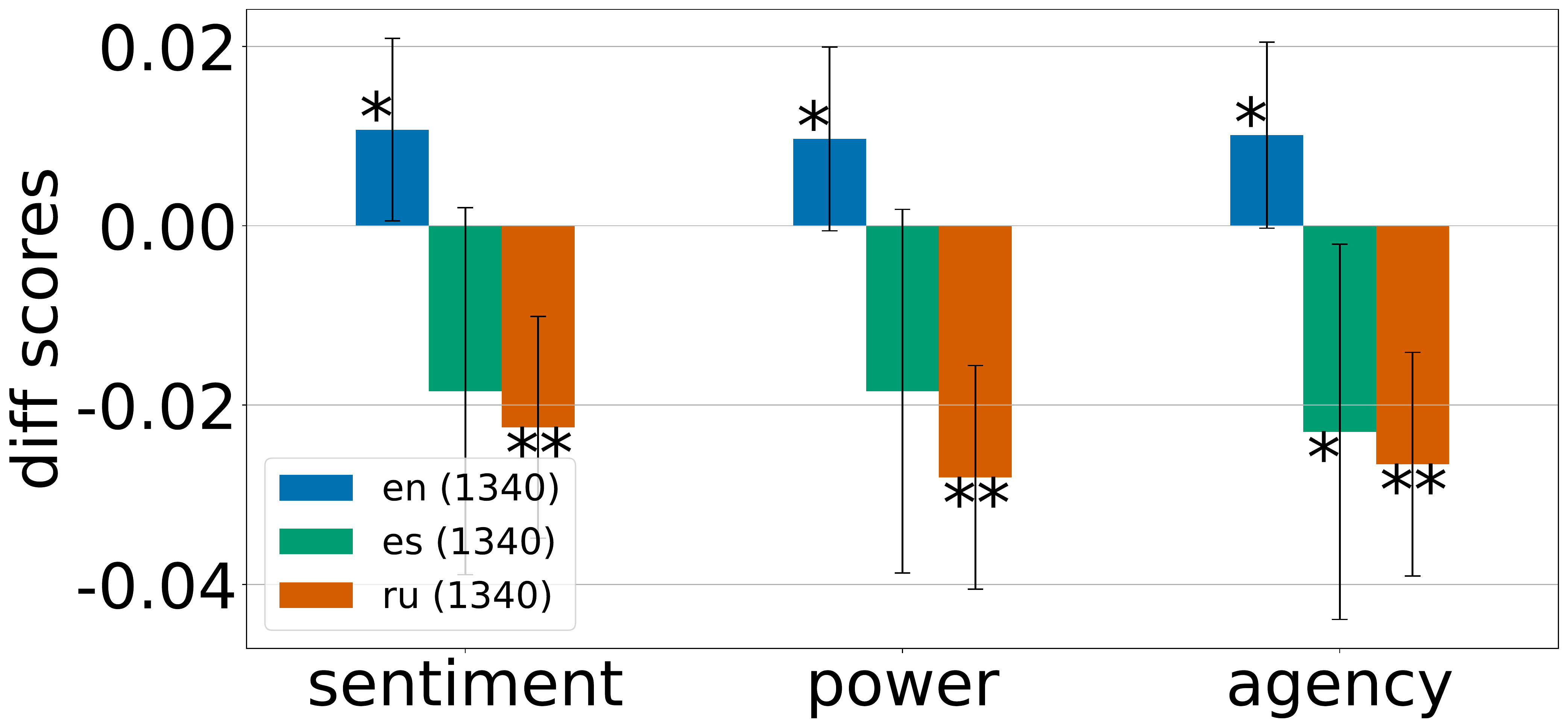}
    \caption{Average differences in affective scores in narratives about LGBT people vs.~matched control people across languages. In Russian and Spanish, LGBT people are consistently portrayed more negatively, with lower power, and with lower agency, whereas in English, they are portrayed more positively. For all figures, asterisks indicate scores are statistically different from zero (paired t-test, $*$:$p{<}0.05$ and $**$:$p{<}0.01$) and brackets denote 0.95 confidence intervals. Numbers in the legend or in the title indicate the number of biographies in each group.}
    \label{fig:overall}
\end{figure}

For each language, we used the best performing model from \Tref{tab:model_augment} to predict contextualized verb annotations for all sentences that contain a target person's name or pronoun as the subject of a verb.\footnote{We omitted Sent(Obj) and focused on Sent(Subj) since sentences where the individual was an object are rare. All subgroups analyzed contained at least 280 verbs.} We then mapped these verb scores to entities (LGBT/matched control people) following \citet{field2019contextual}.
We report \textit{diff score} as the difference between sentiment/power/agency scores in each article about an LGBT person and its matched control, averaged across all pairs (e.g. ``average treatment effect''); a positive score means the articles about LGBT people had a higher affect connotation in aggregate.

\Fref{fig:overall} shows diff scores across the entire corpus. In English, all connotations are significantly positive, whereas in Russian all connotations are significantly negative. All connotations are also negative in Spanish, though only agency is significant. The differences in connotations across languages is surprising, considering that we examine articles about the exact same set of people in all languages.

These trends reflect global perceptions about LGBT people identified by other studies. Data from 2006 and 2011 suggest that many English Wikipedia editors are from the United States, many Russian Wikipedia editors are from Russia, and many Spanish Wikipedia editors are from Spain and (to a lesser extent) other countries, including Argentina, Chile, Netherlands, Mexico, and Venezuela.\footnote{\url{https://meta.wikimedia.org/wiki/Edits_by_project_and_country_of_origin}}\footnote{\url{https://commons.wikimedia.org/w/index.php?title=File:Editor_Survey_Report_-_April_2011.pdf&page=2}} While Wikipedia editor demographics may have since changed, this data shows historical contributions that have been made to Wikipedia and also reflects the countries where these languages are commonly spoken.

While homosexuality was decriminalized in Russia in 1993, homophobia is still prevalent in the country and can manifest, for instance, in laws against ``gay propaganda'' \citep{Wilkinson2014,Buyantueva2018}. A report by the  Wilkins Institute analyzed survey data from 174 countries to measure their social acceptance of LGBT people \citep{Flores2019}. Based on data from 2014-2017, the United States ranked 21\textsuperscript{st} (acceptance score of 7.2), while Russia ranked 120\textsuperscript{th} (score of 3.4). Spain ranked highly (rank: 5, score: 8.1), while other Spanish-speaking countries ranked lower: Argentina (23; 6.9), Chile (27; 6.7), Mexico (32; 6.3), Venezuela (39; 5.7), Peru (53; 5.3). The results in \Fref{fig:overall} suggest that these perceptions are reflected in the \LGBTWikiBio corpus: LGBT people are portrayed with more negative connotations in Russian, but not in English. Results are more mixed in Spanish, which is commonly spoken by (and edited by on Wikipedia) people from a diverse range of countries.

\begin{figure*}[t]
    \centering
    \includegraphics[width=0.95\textwidth]{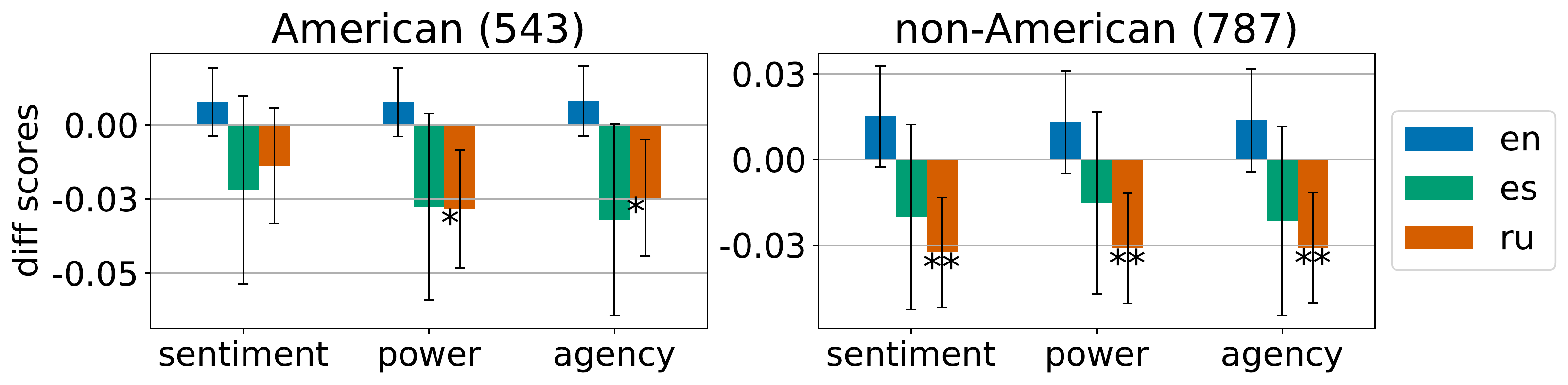}
    \caption{Average differences in affective scores for narratives about nationality subgroups in \LGBTWikiBio.
    }
    \label{fig:nationality}
\end{figure*}

In order to further examine possible cultural differences and in recognition that sexual orientation does not reflect an individual's entire identity, we divide our corpus according to nationality, birth year, and occupation and test if additional social theories manifest in our data.

\Fref{fig:nationality} displays diff scores for each language over American people and non-American people in the corpus. While further subdivisions in nationality could be more informative, we do not report them, out of concern that results over small data sizes could be misleading (e.g.~34 people in our corpus have Russian nationality and 48 have South American nationalities.). In this figure, we investigate the ``local heroes'' hypothesis, which suggests biography pages tend to be longer and more positive for people whose nationality match the language the page is written in \citep{callahan2011cultural,eom2015interactions}. Our data does not show evidence of a local bias, in that trends are nearly identical across articles about American and non-American people, even in English, where we might expect to see different portrayals of Americans and non-Americans.

\begin{figure*}[t]
    \centering
    \includegraphics[width=0.98\textwidth
    ]{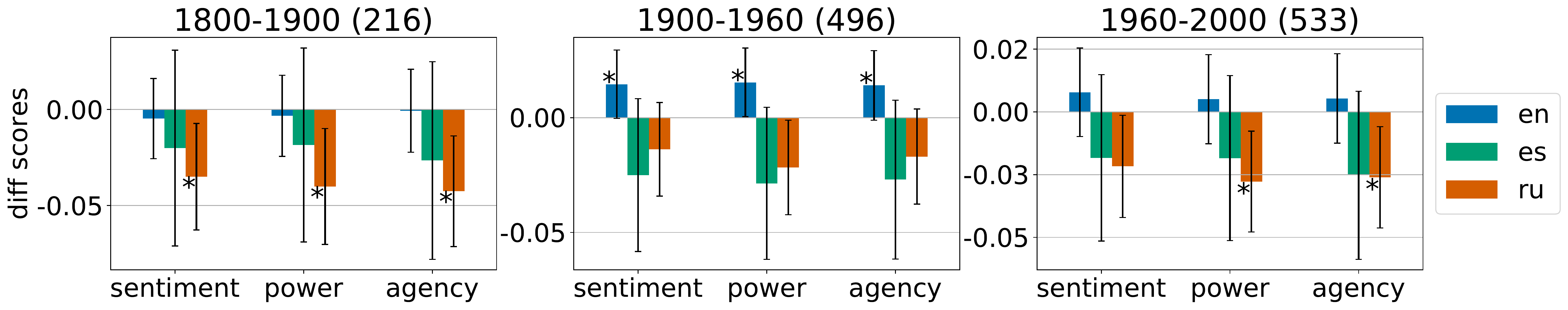}
    \caption{Average sentiment/power/agency diff scores for narratives about age subgroups in \LGBTWikiBio.
    }
    \label{fig:time}
\end{figure*}

In \Fref{fig:time}, we divide our corpus according to the year of birth of each article's subject. Our primary research question in this figure is if the general increase in global acceptance of LGBT people over time is reflected in our data set \citep{Flores2019}. Trends in Russian and Spanish do not significantly change across biography pages for people with different birth years. However, in English, portrayals are neutral/insignificantly negative for people born before 1900, but significantly positive for people born 1900-1960 and (insignificantly) positive for people born after 1960. Thus, while not all results are significant, our data does offer some evidence that changing global perceptions of LGBT people are reflected in their English Wikipedia biography pages.

\begin{figure*}[t]
    \centering
    \includegraphics[width=0.98\textwidth
    ]{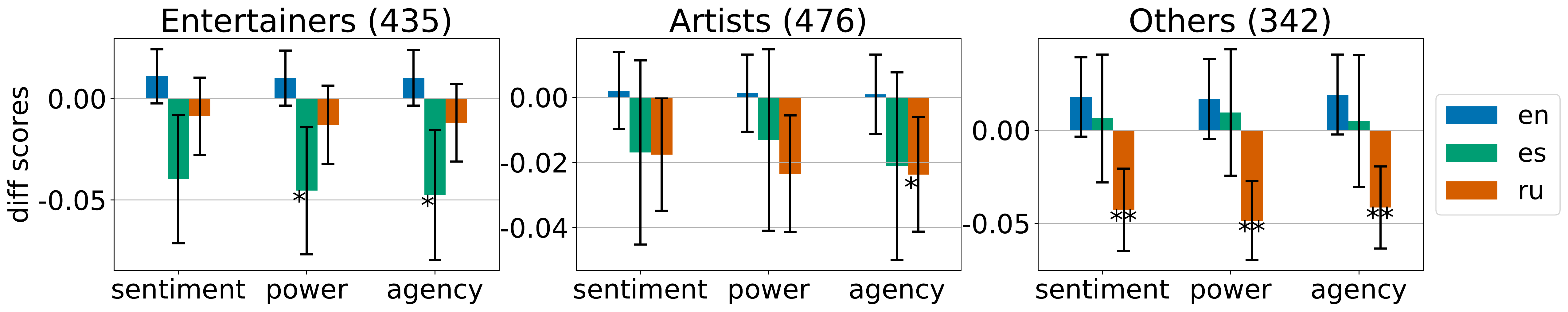}
    \caption{Average sentiment/power/agency diff scores for narratives about occupation subgroups in \LGBTWikiBio.
    }
    \label{fig:occupation}
\end{figure*}

Finally, in \Fref{fig:occupation}, we subdivide the LGBT corpus by occupation. As in \Fref{fig:nationality}, we focus only on the two most common occupations identified in our corpus (Entertainer and Artist) in order to ensure sufficient sample size. Survey and behavioral studies have suggested that LGBT people are perceived as better suited to some occupations than others \cite{Tilcsik2015,Clarke2018}. In \Fref{fig:occupation}, we see little difference in how LGBT people of different occupations are portrayed on Wikipedia in English. However, in Spanish, articles about entertainers have significantly negative connotations, but articles about people of other occupations do not. Conversely, in Russian, while entertainers have near-neutral connotations, people of other occupations, such as politicians, scientists, and activists, are portrayed with significantly negative connotations. While more investigation is needed, our data offers evidence suggesting that perceptions about occupational stereotypes of LGBT people may differ across cultures and languages.

In general, the trends in \Fref{fig:overall} remain consistent across different divisions of the data, with only slight variations. Articles about LGBT people are more negative than controls in Russian, but not in English, and results in Spanish are mixed.

\paragraph{Identification of imbalanced content} One of the intended use-cases of our model is identifying sets of articles that could benefit from manual investigation and possibly revisions. In order to test this use-case, we examined the 10 article pairs that had the greatest difference in power scores between English and Spanish and similarly the 10 pairs with the greatest English-Russian power differentials, where the English article is more positive for both sets. We identify differences in English-Spanish and English-Russian versions of articles about LGBT people, both in what information is included and how it is phrased. We provide three examples drawn from this analysis:

\textit{Cleve Jones}
The Russian article about LGBT activist Cleve Jones spends several sentences describing the AIDS epidemic in the United States including his reaction to it:

\begin{center}
\includegraphics[width=0.9\columnwidth]{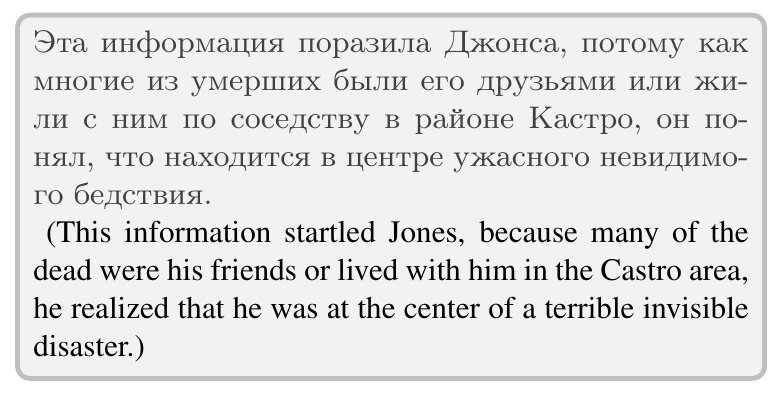}
\end{center}

Our model scores this sentence with negative power and agency and with neutral sentiment, reflecting Jones' role as a passive observer in this particular statement. In contrast, the English version of the article does not include this sentence, nor any detailed descriptions of the U.S. AIDS epidemic. Instead, it focuses on projects that Jones initiated or worked on such as the AIDS Memorial Quilt. The primary difference between language editions in this example is the editors' decision whether to include information that frames Jones more positively or more negatively. Our model also scores this sentence with negative power and agency and with neutral sentiment:

\begin{center}
\includegraphics[width=0.9\columnwidth]{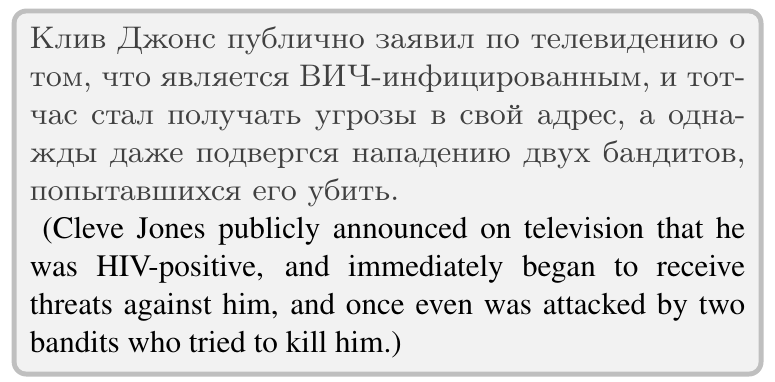}
\end{center}

The English article does say ``Jones described his status as HIV+'' but makes no mention of threats or attacks.

\textit{Hugh Walpole}\footnote{\url{https://en.wikipedia.org/wiki/Hugh_Walpole}} The Russian article about Hugh Walpole, an English novelist, describes:

\begin{center}
\includegraphics[width=0.9\columnwidth]{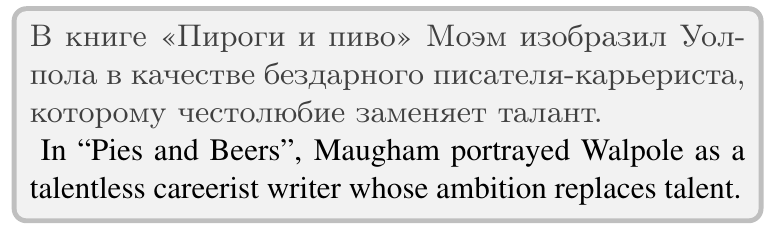}
\end{center}

Our model scores this sentence with negative power and agency, and neutral sentiment. In contrast, the English article states: ``His reputation in literary circles took a blow from a malicious caricature in Somerset Maugham's 1930 novel Cakes and Ale: the character Alroy Kear, a superficial novelist of more pushy ambition than literary talent, was widely taken to be based on Walpole.'' Our model does not score this sentence, as it does not refer to Walpole directly, but rather his reputation and a character based on him. The English article also includes a footnote explaining that Maugham originally denied the character was based on Walpole and only later recanted this denial. The Russian article does not make the distinction between a depiction of Walpole and a character based on Walpole, instead implying that Maugham's critique directly applies to Walpole. The Russian article is also much shorter than the English article, and omits many of his career accomplishments and references to two biographies that portray him positively. For reference, the Spanish article, which is of comparable detail as the Russian article, does not mention Maugham's book.

\textit{Audrey Tang}
The English article about Audrey Tang, a Taiwanese software programmer, describes ``Tang was a child prodigy reading works of classical literature before the age of five...and they began to learn Perl at age 12. Two years later, they dropped out of junior high school, unable to adapt to student life.'' which our model scores with neutral sentiment, power, and agency. The Spanish article states:

\begin{center}
\includegraphics[width=0.9\columnwidth]{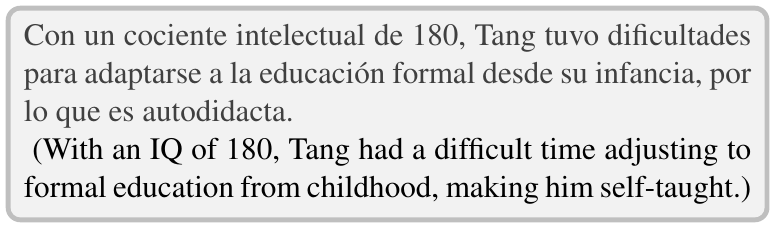}
\end{center}

Our model scores this sentence with negative power, agency, and sentiment, as it focuses on Tang's difficult time and suggests these challenges forced them to drop out, rather than suggesting they did so voluntarily. In general, we found that differences between Spanish and English articles were more subtle than differences between English and Russian articles.

\section{Related Work}
\label{sec:RW}

Our work follows on a series of prior work: \citet{rashkin-etal-2016-connotation} introduced sentiment connotation frames, 
\citet{sap-etal-2017-connotation} extended them to power and agency, and \citet{field2019contextual} introduced the CAA framework. Connotation frames have been used to analyze films, newspaper articles, and online stories \citep{rashkin-etal-2016-connotation,sap-etal-2017-connotation,field2019contextual,antoniak2019narrative}.
\citet{rashkin-etal-2017-multilingual} extend connotation frames to other languages through mapped embeddings, but they do not conduct evaluations against in-language annotations nor provide multilingual annotations.

Our work is generally consistent with existing literature on cross-cultural biases and online biographies.
\citet{Dong:2019} show that perceptions of social roles differ across cultures, while
\citet{De:2019} reveal gender bias in online biographies. 
Other work has examined biases in Wikipedia. \citet{wagner2015s}
show that portrayals of men and women differ across languages, and \citet{callahan2011cultural} reveal systematic cultural biases, particularly in biography pages.

Several studies in social science literature have analyzed biases and their effects on the LGBTQIA+ community, for example, examining mental health \citep{Almeida:2009}
microaggressions \citep{Balsam:2011},
and sociopolitical involvement \citep{men:2013}.
With a few exceptions \citep{schmidt:2017:survey,Fast:2016,Dinakar:2012,mendelsohn2020framework}, biased language about or against the LGBTQIA+ community has not been examined and analyzed extensively in automated analyses.
The closest study to ours is an examination of gender, race, and LGBT portrayals in $700$ popular films
\cite{Smith:2015}.

\section{Conclusion}
Our work provides methodology and datasets that extend the capabilities of affective analysis to multilingual settings.
While we focus on Wikipedia, our methodology could be used to conduct analyses in any English, Russian, and Spanish narrative text, which can aid writers in obtaining a neutral point of view and provide insight into social stereotypes, especially when used in combination with other methods.
This framework supports the investigation of a wide range of research questions, and offers multiple avenues for future work such as improving the multilingual model, expansion to additional languages, investigation of Wikipedia edit histories, and the incorporation of additional connotations and existing linguistic databases.

\section*{Acknowledgements}

We gratefully acknowledge Alan Black, Artidoro Pagnoni, Maria Ryskina, Santiago Castro, Luis Figueroa, Shuly Wintner and our anonymous reviewers for providing feedback and aiding with translations. AF gratefully acknowledges support from the Google PhD Fellowship. This material is based upon work supported by the National Science Foundation (NSF) Graduate Research Fellowship Program under Grant No.~DGE1745016 and by NSF Grants No.~IIS1812327 and IIS2040926. Any opinions, findings, and conclusions or recommendations expressed in this material are those of the authors and do not necessarily reflect the views of the NSF.

\appendix
\setcounter{page}{1}
\section{Multilingual Annotation Collection}
\label{sec:schema}
We provide additional details of data collection to facilitate reproducibility and fully describe data quality.

\noindent\textbf{Language Choice} All annotations were crowdsourced through the Figure~8 platform.\footnote{Figure~8 is now called Appen (\url{https://appen.com/})} Our original target languages for this task were English, Russian, Mandarin Chinese, and French. We constructed three rounds of in-house pilot studies for English and one round for Chinese before we released the annotation tasks. After releasing these tasks, we received almost no annotations in French or Chinese, despite increasing payment, expanding the number of target countries, and relaunching tasks. We ultimately dropped Chinese and French in favor of Spanish, for which we were able to obtain annotations. 

\noindent\textbf{Instructions}
Task instructions were originally written in English and then translated to other languages by native speakers. For each language, a second native speaker checked the translation. The same native speakers also examined the data samples to be annotated, in order to ensure that contexts were grammatical and representative. Heuristics for generating samples were revised according to their feedback.

While power and sentiment are generally well-known terms, agency is unfamiliar to most people and can be difficult to define. Additionally, our Russian, Chinese, and French translators determined that there is no single-word translation for ``agency'' in these languages, and they instead used combinations of other words to define the concept. Thus, in the annotation task, following \citet{sap-etal-2017-connotation} we provided three agency ``priming questions'' to the annotators.

\noindent\textbf{Task Settings}
We placed several restrictions on the pool of annotators in order to ensure annotation quality.

Each annotation task included five to eight examples in the task instructions, which were then turned into ``quiz questions''. Annotators needed to answer an initial eight quiz questions with $70\%$ accuracy to begin the task. As the task proceeded, an additional $10$ quiz questions were interspersed with examples to be annotated, and annotators needed to maintain a $70\%$ score on these questions in order to continue the task. We made our questions extremely similar to the examples given in the task instructions, because affective connotations can be subjective, and it is difficult to construct quiz questions that we can expect all high-quality annotators to consistently answer correctly. Instead, our quiz questions ensured that annotators read and understood the instructions.

We released data in batches of $100$-$200$ annotation examples, as we found that larger batch sizes did not complete. Small batches also allowed us to block annotators who failed quiz questions in earlier batches from attempting to complete later batches. Additionally, we disabled the chrome translation plugin for all non-English tasks.

For Spanish, we restricted annotators to people from nine South America countries: Argentina, Bolivia, Chile, Colombia, Ecuador, Paraguay, Peru, Uruguay and Venezuela. We restricted English annotators to the United States and Russian annotators to Russia. Payment for annotation tasks was set based on the time taken to complete the task during pilot studies and the minimum wage of target countries. We also adjusted pay based on survey feedback from early batches. The final rates are in \Tref{tab:payrate}. In general, we paid agency substantially higher than the other two tasks because we had three additional priming questions that annotators needed to annotate for each instance. 
\begin{table}[h]
    \centering
    \begin{tabular}{*4c}
    Task & English & Russian & Spanish \\
    \hline \hline
    \textbf{Power} & 20 & 4 & 5\\
    \textbf{Agency} & 40 & 8 & 8\\
    \textbf{Sent} & 20 & 6 & 5\\
    \end{tabular}
    \caption{Task pay rates in cents per five instances}
    \label{tab:payrate}
\end{table}

Finally, as mentioned in \Sref{sec:data}, we screened out annotations from lower-quality annotators after data collection. For each annotator, we computed how often that annotator judged an instance differently than the other two annotators who judged that instance. We then removed annotations from any annotators whose disagreement rate was greater than one standard deviation away from the mean disagreement rate.
In our final dataset, we keep only instances that have at least two judgements after removing these annotators.
\Tref{tab:agreement} shows the full agreement scores for each annotation task after post-processing and \Tref{tab:statistics} shows the number of annotated instances for each language.

\begin{table}[h]
    \centering
        \begin{tabular}{ccccc}
    &  English & Russian & Spanish \\ \hline \hline
    \textbf{Power} & 0.27 & 0.33 & 0.25 \\
    \textbf{Agency} & 0.20 & 0.23 & 0.24 \\
    \textbf{Sent(subj)} & 0.20 & 0.27 & 0.22 \\
    \textbf{Sent(obj)} & 0.22 & 0.39 & 0.31 \\
    \end{tabular}
    
    \caption{Krip.'s Alpha per task, after post-processing.}
    \label{tab:agreement}
\end{table}

\begin{table}[h]
    \centering
        \begin{tabular}{ccccc}
    &  English & Russian & Spanish \\ \hline \hline
    \textbf{Power} & 837 & 880  & 877\\
    \textbf{Agency} & 888 & 879 & 888 \\
    \textbf{Sent(subj)} & 860 & 868 & 808 \\
    \textbf{Sent(obj)} & 860 & 868  & 808\
    \end{tabular}
    
    \caption{Num. of annotated instances, after post-processing.}
    \label{tab:statistics}
\end{table}

\section{Model Details}
\label{sec:appendix_classifier}

Each logistic regression classification model has 3,075 parameters. 
We did a grid search over the weights given for each class and chose the final weights based on the validation set F1 score. 
We release all trained models and their hyperparameter configuration as a part of our codebase.

\Tref{tab:model_baseline} validates that the performance of our XLM-based model is comparable to prior work by evaluating our model on the same data used in \citet{rashkin-etal-2016-connotation} and \citet{field2019contextual}. In this setting, where we evaluate our model on uncontextualized annotations, we decontextualize the XLM embeddings in the same way as \citet{field2019contextual}. Thus, the primary difference between our model and theirs is the use of cross-lingual XLM embeddings instead of ELMo embeddings.
Our model performs similarly with \citet{rashkin-etal-2016-connotation}. Although our model is slightly worse than \citet{field2019contextual}, this drop is not surprising and considered a cost of making our model language-agnostic.

\begin{table}[h]
\resizebox{\columnwidth}{!}{
\begin{tabular}{@{}lcccc@{}}
% \toprule
& \textbf{Sent\textsubscript{subj}} & \textbf{Sent\textsubscript{obj}} & \textbf{Power} & \textbf{Agency} \\  \hline \hline
Majority baseline&  26.2    & 28.7       & 27.4            & 29.5        \\
\citet{rashkin-etal-2016-connotation}    & 66.6          & 37.4 & 51.8           & 46.5           \\
\citet{field2019contextual}    & 61.1   & 40.4      & 56.0       & 48.8        \\
Our model            & 54.6  & 45.0         & 47.4            & 45.0       \\ 
\end{tabular}
}
\caption{Classification evaluation results in macro F1 score. Majority baseline always outputs the most frequent label in the data. Numbers in the \citet{field2019contextual} row are directly borrowed from the original paper. 
}
\label{tab:model_baseline}
\end{table}

\section{\LGBTWikiBio Corpus Construction}
\label{sec:appendix_matching}

As described in \Sref{sec:LGBTCorp}, we collected Wikipedia biography pages about LGBT people using lists of LGBT people from English\footnote{\url{https://en.wikipedia.org/wiki/List_of_gay,_lesbian_or_bisexual_people} \\\url{https://en.wikipedia.org/wiki/List_of_transgender_people}} and Spanish\footnote{\url{https://es.wikipedia.org/wiki/Anexo:Famosos_que_han_salido_del_armario}} Wikipedia. We additionally include people with Wikidata property P91 values of Q6636 (homosexuality), Q6649 (lesbian), Q43200 (bisexuality), or Q592 (gay), and people with Wikidata property P21 of Q1052281 (transgender female) or Q2449503 (transgender male). We removed people who do not have pages in all target languages (English, Spanish, Russian) and who have $<3$ sentences to be analyzed in any language. We define sentences to be analyzed as ones containing the person's name or pronoun as a subject or object. We automatically inferred pronouns based on whether ``he'' or ``she'' is more frequent in the article text, which we expect to be effective even for transgender people because of Wikipedia's Manual of Style/Gender identity.\footnote{\url{https://en.wikipedia.org/wiki/Wikipedia:Manual_of_Style/Gender_identity}}

After filtering, the LGBT-half of the \LGBTWikiBio corpus contains $1,340$ biography pages.

We identify matched control biography pages using the algorithm from \citet{anonymous20matching}. We apply the same filters to candidate control pages as to LGBT pages (discarding articles with $<3$ sentences to be analyzed in English, Russian, or Spanish). The matching algorithm using TF-IDF vectors constructed from biography page categories as matching features. The vectors contain a pivot-slope correction term, which is intended to prevent the method from favoring pages with fewer categories \cite{singhal2017pivoted}. Following the recommendations in \citet{anonymous20matching}, we set the pivot to the average number of categories per article in our data set. We then tune the slope until the LGBT-half and the matched controls have the same number of average categories (excluding LGBT-specific categories). We try pivot values in 0.1 increments between [0, 0.5], and fix the pivot as 0.1. In \Tref{Sample List} we provide examples of constructed pairs.

\begin{table}[h]
\centering
\resizebox{0.97\columnwidth}{!}{
\begin{tabular}{ll}
\toprule
LGBT (Non-LGBT pair) & Common Categories (sampled three) \\
\midrule
Tim Cook (Steve Jobs)  &  \pbox{7cm}{\textit{Apple\_Inc.\_executives} \\ \textit{American\_computer\_businesspeople}\\\textit{21st-century\_American\_businesspeople}}\\
\midrule
Plato (Aristotle)  &  \pbox{7cm}{\textit{4th-century\_BC\_philosophers} \\ \textit{Ancient\_Greek\_political\_philosophers}\\\textit{4th-century\_BC\_writers}}\\
\midrule
Lily Allen (Dua Lipa) & \pbox{7cm}{\textit{English\_female\_singer-songwriters}\\\textit{Brit\_Award\_winners} \\ \textit{Electropop\_musicians}}\\
\midrule
Tom Ford (Anna Sui) & \pbox{7cm}{\textit{American\_fashion\_businesspeople}\\\textit{Luxury\_brands} \\ \textit{Parsons\_School\_of\_Design\_alumni}}\\
\bottomrule
\end{tabular}
}
\caption{A sampled list of paired-people from \LGBTWikiBio.}
\label{Sample List}
\end{table}

\fontsize{9pt}{10.2pt} \selectfont
\bibliography{aaai21}

\begin{thebibliography}{42}
\providecommand{\natexlab}[1]{#1}
\providecommand{\url}[1]{\texttt{#1}}
\providecommand{\urlprefix}{URL }
\expandafter\ifx\csname urlstyle\endcsname\relax
  \providecommand{\doi}[1]{doi:\discretionary{}{}{}#1}\else
  \providecommand{\doi}{doi:\discretionary{}{}{}\begingroup
  \urlstyle{rm}\Url}\fi

\bibitem[{Almeida et~al.(2009)Almeida, Johnson, Corliss, Molnar, and
  Azrael}]{Almeida:2009}
Almeida, J.; Johnson, R.~M.; Corliss, H.~L.; Molnar, B.~E.; and Azrael, D.
  2009.
\newblock Emotional Distress Among LGBT Youth: The Influence of Perceived
  Discrimination Based on Sexual Orientation.
\newblock \emph{Journal of Youth and Adolescence} 38(7): 1001--1014.
\newblock ISSN 1573-6601.
\newblock \doi{10.1007/s10964-009-9397-9}.

\bibitem[{Antoniak, Mimno, and Levy(2019)}]{antoniak2019narrative}
Antoniak, M.; Mimno, D.; and Levy, K. 2019.
\newblock Narrative Paths and Negotiation of Power in Birth Stories.
\newblock In \emph{Proc. ACM Hum.-Comput. Interact.}, volume~3, 88. ACM.

\bibitem[{Balsam et~al.(2011)Balsam, Molina, Beadnell, Simoni, and
  Walters}]{Balsam:2011}
Balsam, K.~F.; Molina, Y.; Beadnell, B.; Simoni, J.; and Walters, K. 2011.
\newblock Measuring Multiple Minority Stress: The LGBT People of Color
  Microaggressions Scale.
\newblock \emph{Cultur Divers Ethnic Minor Psychol.} 17(2): 163--174.

\bibitem[{Bamman, O{'}Connor, and Smith(2013)}]{bamman-etal-2013-learning}
Bamman, D.; O{'}Connor, B.; and Smith, N.~A. 2013.
\newblock Learning Latent Personas of Film Characters.
\newblock In \emph{Proc. of ACL}, 352--361. Sofia, Bulgaria: Association for
  Computational Linguistics.

\bibitem[{Barrault et~al.(2019)Barrault, Bojar, Costa-juss{\`a}, Federmann,
  Fishel, Graham, Haddow, Huck, Koehn, Malmasi, Monz, M{\"u}ller, Pal, Post,
  and Zampieri}]{barrault-EtAl:2019:WMT}
Barrault, L.; Bojar, O.; Costa-juss{\`a}, M.~R.; Federmann, C.; Fishel, M.;
  Graham, Y.; Haddow, B.; Huck, M.; Koehn, P.; Malmasi, S.; Monz, C.;
  M{\"u}ller, M.; Pal, S.; Post, M.; and Zampieri, M. 2019.
\newblock Findings of the 2019 Conference on Machine Translation ({WMT}19).
\newblock In \emph{Proc. of WMT}, 1--61. Florence, Italy: Association for
  Computational Linguistics.
\newblock \doi{10.18653/v1/W19-5301}.

\bibitem[{Buyantueva(2018)}]{Buyantueva2018}
Buyantueva, R. 2018.
\newblock LGBT Rights Activism and Homophobia in Russia.
\newblock \emph{Journal of Homosexuality} 65(4): 456--483.
\newblock \doi{10.1080/00918369.2017.1320167}.
\newblock \urlprefix\url{https://doi.org/10.1080/00918369.2017.1320167}.
\newblock PMID: 28409697.

\bibitem[{Callahan and Herring(2011)}]{callahan2011cultural}
Callahan, E.~S.; and Herring, S.~C. 2011.
\newblock Cultural bias in Wikipedia content on famous persons.
\newblock \emph{Journal of the American society for information science and
  technology} 62(10): 1899--1915.

\bibitem[{Chandrasekharan et~al.(2017)Chandrasekharan, Pavalanathan,
  Srinivasan, Glynn, Eisenstein, and Gilbert}]{Chandrasekharan:2017}
Chandrasekharan, E.; Pavalanathan, U.; Srinivasan, A.; Glynn, A.; Eisenstein,
  J.; and Gilbert, E. 2017.
\newblock You can't stay here: The efficacy of reddit's 2015 ban examined
  through hate speech.
\newblock \emph{Proc. ACM Hum.-Comput. Interact.} 1(CSCW): 31.

\bibitem[{Clarke and Arnold(2018)}]{Clarke2018}
Clarke, H.~M.; and Arnold, K.~A. 2018.
\newblock The Influence of Sexual Orientation on the Perceived Fit of Male
  Applicants for Both Male- and Female-Typed Jobs.
\newblock \emph{Frontiers in Psychology} 9: 656.
\newblock \doi{10.3389/fpsyg.2018.00656}.

\bibitem[{Conneau and Lample(2019)}]{conneau2019cross}
Conneau, A.; and Lample, G. 2019.
\newblock Cross-lingual Language Model Pretraining.
\newblock In \emph{Advances in Neural Information Processing Systems},
  7057--7067.

\bibitem[{De-Arteaga et~al.(2019)De-Arteaga, Romanov, Wallach, Chayes, Borgs,
  Chouldechova, Geyik, Kenthapadi, and Kalai}]{De:2019}
De-Arteaga, M.; Romanov, A.; Wallach, H.; Chayes, J.; Borgs, C.; Chouldechova,
  A.; Geyik, S.; Kenthapadi, K.; and Kalai, A.~T. 2019.
\newblock Bias in bios: A case study of semantic representation bias in a
  high-stakes setting.
\newblock In \emph{Proc. of FAT}, 120--128. ACM.

\bibitem[{Dinakar et~al.(2012)Dinakar, Jones, Havasi, Lieberman, and
  Picard}]{Dinakar:2012}
Dinakar, K.; Jones, B.; Havasi, C.; Lieberman, H.; and Picard, R. 2012.
\newblock Common sense reasoning for detection, prevention, and mitigation of
  cyberbullying.
\newblock \emph{ACM Transactions on Interactive Intelligent Systems (TiiS)}
  2(3): 18.

\bibitem[{Doi and Stewart(2019)}]{doi_stewart_2019}
Doi, K.; and Stewart, P. 2019.
\newblock Interview: The Invisible Struggle of Japan’s Transgender
  Population.
\newblock \emph{Human Rights Watch} .

\bibitem[{Dong et~al.(2019)Dong, Jurgens, Banea, and Mihalcea}]{Dong:2019}
Dong, M.; Jurgens, D.; Banea, C.; and Mihalcea, R. 2019.
\newblock Perceptions of Social Roles Across Cultures.
\newblock In \emph{International Conference on Social Informatics}, 157--172.
  Springer.

\bibitem[{Eckert(2000)}]{eckert2000language}
Eckert, P. 2000.
\newblock \emph{Language variation as social practice: The linguistic
  construction of identity in Belten High}.
\newblock Wiley-Blackwell.

\bibitem[{Eom et~al.(2015)Eom, Arag{\'o}n, Laniado, Kaltenbrunner, Vigna, and
  Shepelyansky}]{eom2015interactions}
Eom, Y.-H.; Arag{\'o}n, P.; Laniado, D.; Kaltenbrunner, A.; Vigna, S.; and
  Shepelyansky, D.~L. 2015.
\newblock Interactions of cultures and top people of Wikipedia from ranking of
  24 language editions.
\newblock \emph{PloS one} 10(3).

\bibitem[{Fast and Horvitz(2016)}]{Fast:2016}
Fast, E.; and Horvitz, E. 2016.
\newblock Identifying Dogmatism in Social Media: Signals and Models.
\newblock In \emph{Proc. of EMNLP}, 690--699. Austin, Texas: Association for
  Computational Linguistics.
\newblock \doi{10.18653/v1/D16-1066}.

\bibitem[{Fellbaum(2012)}]{fellbaum2012wordnet}
Fellbaum, C. 2012.
\newblock WordNet.
\newblock \emph{The encyclopedia of applied linguistics} .

\bibitem[{Field, Bhat, and Tsvetkov(2019)}]{field2019contextual}
Field, A.; Bhat, G.; and Tsvetkov, Y. 2019.
\newblock Contextual Affective Analysis: A Case Study of People Portrayals in
  Online \#MeToo Stories.
\newblock In \emph{Proc. of ICSWM 2019}.

\bibitem[{Field, Park, and Tsvetkov(2020)}]{anonymous20matching}
Field, A.; Park, C.~Y.; and Tsvetkov, Y. 2020.
\newblock Controlled Analyses of Social Biases in Wikipedia Bios.
\newblock ArXiv preprint arXiv:2101.00078.

\bibitem[{Field and Tsvetkov(2019)}]{field-tsvetkov-2019-entity}
Field, A.; and Tsvetkov, Y. 2019.
\newblock Entity-Centric Contextual Affective Analysis.
\newblock In \emph{Proc. of ACL}, 2550--2560. Florence, Italy: Association for
  Computational Linguistics.
\newblock \doi{10.18653/v1/P19-1243}.

\bibitem[{Flores(2019)}]{Flores2019}
Flores, A.~R. 2019.
\newblock Social Acceptance of LGBT People in 174 Countries.
\newblock
  \urlprefix\url{https://williamsinstitute.law.ucla.edu/publications/global-acceptance-index-lgbt/}.

\bibitem[{Fournier, Moskowitz, and Zuroff(2002)}]{fournier2002social}
Fournier, M.~A.; Moskowitz, D.; and Zuroff, D.~C. 2002.
\newblock Social rank strategies in hierarchical relationships.
\newblock \emph{Journal of Personality and Social Psychology} 83(2): 425.

\bibitem[{Hall and Braunwald(1981)}]{Hall:Braunwald:1981}
Hall, J.~A.; and Braunwald, K.~G. 1981.
\newblock Gender Cues in Conversations.
\newblock \emph{Journal of Personality and Social Psychology} 40(1): 99.

\bibitem[{Harris et~al.(2013)Harris, Battle, Pastrana, and Daniels}]{men:2013}
Harris, A.; Battle, J.; Pastrana, Antonio~(., J.; and Daniels, J. 2013.
\newblock The Sociopolitical Involvement of Black, Latino, and Asian/Pacific
  Islander Gay and Bisexual Men.
\newblock \emph{Journal of Men's Studies} 21(3): 236--254.

\bibitem[{Lisitza(2017)}]{lisitza_2017}
Lisitza, A. 2017.
\newblock History of Pride Parades in the U.S.
\newblock \emph{TeenVogue} .

\bibitem[{Mendelsohn, Tsvetkov, and Jurafsky(2020)}]{mendelsohn2020framework}
Mendelsohn, J.; Tsvetkov, Y.; and Jurafsky, D. 2020.
\newblock A Framework for the Computational Linguistic Analysis of
  Dehumanization.
\newblock \emph{Front. Artif. Intell.} \doi{10.3389/frai.2020.00055}.

\bibitem[{Miller(1995)}]{Miller:1995:WLD:219717.219748}
Miller, G.~A. 1995.
\newblock WordNet: A Lexical Database for English.
\newblock \emph{Commun. ACM} 38(11): 39--41.
\newblock ISSN 0001-0782.
\newblock \doi{10.1145/219717.219748}.

\bibitem[{Mohammad(2018)}]{mohammad-2018-obtaining}
Mohammad, S. 2018.
\newblock Obtaining Reliable Human Ratings of Valence, Arousal, and Dominance
  for 20,000 {E}nglish Words.
\newblock In \emph{Proc. of ACL}, 174--184. Melbourne, Australia: Association
  for Computational Linguistics.
\newblock \doi{10.18653/v1/P18-1017}.

\bibitem[{Mohammad and Turney(2013)}]{Mohammad13}
Mohammad, S.~M.; and Turney, P.~D. 2013.
\newblock Crowdsourcing a Word-Emotion Association Lexicon.
\newblock \emph{Computational Intelligence} 29(3): 436--465.

\bibitem[{Rashkin et~al.(2017)Rashkin, Bell, Choi, and
  Volkova}]{rashkin-etal-2017-multilingual}
Rashkin, H.; Bell, E.; Choi, Y.; and Volkova, S. 2017.
\newblock Multilingual Connotation Frames: A Case Study on Social Media for
  Targeted Sentiment Analysis and Forecast.
\newblock In \emph{Proc. of ACL}, 459--464. Vancouver, Canada: Association for
  Computational Linguistics.
\newblock \doi{10.18653/v1/P17-2073}.

\bibitem[{Rashkin, Singh, and Choi(2016)}]{rashkin-etal-2016-connotation}
Rashkin, H.; Singh, S.; and Choi, Y. 2016.
\newblock Connotation Frames: A Data-Driven Investigation.
\newblock In \emph{Proc. of ACL}, 311--321. Berlin, Germany: Association for
  Computational Linguistics.
\newblock \doi{10.18653/v1/P16-1030}.

\bibitem[{Recasens, Danescu-Niculescu-Mizil, and
  Jurafsky(2013)}]{recasens-etal-2013-linguistic}
Recasens, M.; Danescu-Niculescu-Mizil, C.; and Jurafsky, D. 2013.
\newblock Linguistic Models for Analyzing and Detecting Biased Language.
\newblock In \emph{Proc. of ACL}, 1650--1659. Sofia, Bulgaria: Association for
  Computational Linguistics.

\bibitem[{Sap et~al.(2017)Sap, Prasettio, Holtzman, Rashkin, and
  Choi}]{sap-etal-2017-connotation}
Sap, M.; Prasettio, M.~C.; Holtzman, A.; Rashkin, H.; and Choi, Y. 2017.
\newblock Connotation Frames of Power and Agency in Modern Films.
\newblock In \emph{Proc. of EMNLP}, 2329--2334. Copenhagen, Denmark:
  Association for Computational Linguistics.
\newblock \doi{10.18653/v1/D17-1247}.

\bibitem[{Schmidt and Wiegand(2017)}]{schmidt:2017:survey}
Schmidt, A.; and Wiegand, M. 2017.
\newblock A Survey on Hate Speech Detection using Natural Language Processing.
\newblock In \emph{Proc. of SocialNLP}, 1--10. Valencia, Spain: Association for
  Computational Linguistics.
\newblock \doi{10.18653/v1/W17-1101}.

\bibitem[{Singhal, Buckley, and Mitra(1996)}]{singhal2017pivoted}
Singhal, A.; Buckley, C.; and Mitra, M. 1996.
\newblock Pivoted document length normalization.
\newblock In \emph{Proc. of SIGIR}, volume~51, 176--184. ACM New York, NY, USA.

\bibitem[{Smith et~al.(2015)Smith, Choueiti, Pieper, Gillig, Lee, and
  DeLuca}]{Smith:2015}
Smith, S.; Choueiti, M.; Pieper, K.; Gillig, T.; Lee, C.; and DeLuca, D. 2015.
\newblock Inequality in 700 Popular Films: Examining Portrayals of Gender,
  Race, \& {LGBT} Status from 2007 to 2014.
\newblock \emph{Institute for Diversity and Empowerment at Annenberg} .

\bibitem[{Stuart(2010)}]{Stuart2010}
Stuart, E. 2010.
\newblock Matching methods for causal inference: A review and a look forward.
\newblock \emph{Stat Sci} 25(1): 1--21.
\newblock \doi{doi:10.1214/09-STS313}.

\bibitem[{Tannen(1994)}]{tannen1994gender}
Tannen, D. 1994.
\newblock \emph{Gender and discourse}.
\newblock Oxford University Press.

\bibitem[{Tilcsik, Anteby, and Knight(2015)}]{Tilcsik2015}
Tilcsik, A.; Anteby, M.; and Knight, C.~R. 2015.
\newblock Concealable Stigma and Occupational Segregation: Toward a Theory of
  Gay and Lesbian Occupations.
\newblock \emph{Administrative Science Quarterly} 60(3): 446--481.
\newblock \doi{10.1177/0001839215576401}.
\newblock \urlprefix\url{https://doi.org/10.1177/0001839215576401}.

\bibitem[{Wagner et~al.(2015)Wagner, Garcia, Jadidi, and
  Strohmaier}]{wagner2015s}
Wagner, C.; Garcia, D.; Jadidi, M.; and Strohmaier, M. 2015.
\newblock It's a man's {W}ikipedia? {A}ssessing gender inequality in an online
  encyclopedia.
\newblock In \emph{Proc. of ICWSM}.

\bibitem[{Wilkinson(2014)}]{Wilkinson2014}
Wilkinson, C. 2014.
\newblock Putting “Traditional Values” Into Practice: The Rise and
  Contestation of Anti-Homopropaganda Laws in Russia.
\newblock \emph{Journal of Human Rights} 13(3): 363--379.
\newblock \doi{10.1080/14754835.2014.919218}.
\newblock \urlprefix\url{https://doi.org/10.1080/14754835.2014.919218}.

\end{thebibliography}

\end{document}